\documentclass[conference]{IEEEtran}
\IEEEoverridecommandlockouts
\usepackage{cite}
\usepackage{amsmath,amssymb,amsfonts}
\usepackage{algorithmic}
\usepackage{graphicx}
\usepackage{textcomp}
\usepackage{xcolor}
\usepackage{booktabs} 
\usepackage{pifont} 
\usepackage{url}
\usepackage{adjustbox}

\usepackage{multirow}
\usepackage{colortbl}

\newcommand{\etal}{\textit{et al}. }
\newcommand{\ie}{\textit{i}.\textit{e}., }

\def\BibTeX{{\rm B\kern-.05em{\sc i\kern-.025em b}\kern-.08em
    T\kern-.1667em\lower.7ex\hbox{E}\kern-.125emX}}
\begin{document}

\title{Semantics Preserving Emoji Recommendation with Large Language Models
}

\makeatletter
\newcommand{\linebreakand}{%
  \end{@IEEEauthorhalign}
  \hfill\mbox{}\par
  \mbox{}\hfill\begin{@IEEEauthorhalign}
}
\makeatother

\author{\IEEEauthorblockN{Zhongyi Qiu}
\IEEEauthorblockA{\textit{School of Computational Science and Engineering} \\
\textit{Georgia Institute of Technology}\\
Atlanta, GA, USA \\
zhongyiqiu@gatech.edu}
\and
\IEEEauthorblockN{Kangyi Qiu}
\IEEEauthorblockA{\textit{School of Computer Science} \\
\textit{Georgia Institute of Technology}\\
Atlanta, GA, USA \\
kqiu37@gatech.edu}
\and
\IEEEauthorblockN{Hanjia Lyu}
\IEEEauthorblockA{\textit{Department of Computer Science} \\
\textit{University of Rochester}\\
Rochester, NY, USA \\
hlyu5@ur.rochester.edu}
\linebreakand
\IEEEauthorblockN{Wei Xiong}
\IEEEauthorblockA{\textit{Department of Computer Science} \\
\textit{University of Rochester}\\
Rochester, NY, USA \\
wxiongur@gmail.com}
\and
\IEEEauthorblockN{Jiebo Luo}
\IEEEauthorblockA{\textit{Department of Computer Science} \\
\textit{University of Rochester}\\
Rochester, NY, USA \\
jluo@cs.rochester.edu}
}

\maketitle

\begin{abstract}
Emojis have become an integral part of digital communication, enriching text by conveying emotions, tone, and intent. Existing emoji recommendation methods are primarily evaluated based on their ability to match the exact emoji a user chooses in the original text. However, they ignore the essence of users' behavior on social media in that each text can correspond to multiple reasonable emojis. To better assess a model's ability to align with such real-world emoji usage, we propose a new semantics preserving evaluation framework for emoji recommendation, which measures a model's ability to recommend emojis that maintain the semantic consistency with the user's text. To evaluate how well a model preserves semantics, we assess whether the predicted affective state, demographic profile, and attitudinal stance of the user remain unchanged. If these attributes are preserved, we consider the recommended emojis to have maintained the original semantics. 
The advanced abilities of Large Language Models (LLMs) in understanding and generating nuanced, contextually relevant output make them well-suited for handling the complexities of semantics preserving emoji recommendation. 
To this end, we construct a comprehensive benchmark to systematically assess the performance of six proprietary and open-source LLMs using different prompting techniques on our task.  Our experiments demonstrate that GPT-4o outperforms other LLMs, achieving a semantics preservation score of 79.23\%. Additionally, we conduct case studies to analyze model biases in downstream classification tasks and evaluate the diversity of the recommended emojis.
\end{abstract}

\begin{IEEEkeywords}
semantics preserving, emoji recommendation, large language models, prompt engineering
\end{IEEEkeywords}

\section{Introduction}

Emojis play a key role in modern digital communication, offering a simple way to convey emotions, tone, and intent that text alone may not fully capture. They are widely used across various platforms, including social media, instant messaging, and even professional digital communications, contributing to more expressive textual experience~\cite{telaumbanua2024use}. However, with the vast number of emojis available, it can be challenging to users to select the most appropriate one to convey their intended message~\cite{zhao2020caper}. To better assist users in selecting appropriate emojis, emoji recommendation tasks have emerged, aiming to provide models with an input text and relevant context to recommend the most contextually appropriate emojis~\cite{barbieri2018semeval}.

\begin{figure}
    \centering
    \includegraphics[width=\linewidth]{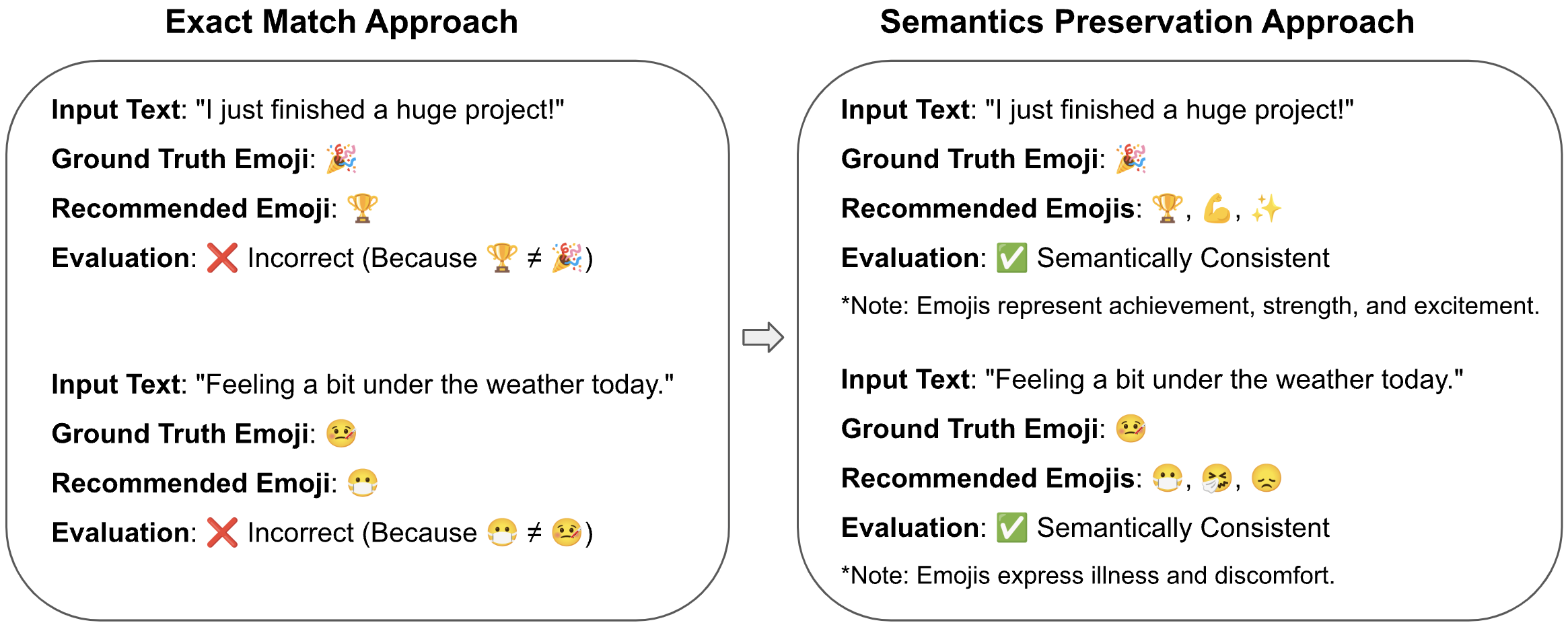}
    \caption{Comparison of traditional exact match and semantics preservation approaches for emoji recommendation. The semantics preservation approach can suggest multiple emojis that maintain the semantic meaning of the text, even if they differ from the ground truth emojis.}
    \label{fig:semantic_preservation}
\end{figure}

Existing emoji recommendation methods are primarily evaluated through exact match, where the recommended emoji is compared to the one the user originally selected~\cite{zhao2018analyzing,barbieri2018multimodal,peng2021seq2emoji}. This evaluation criterion, while straightforward, introduces several limitations. 
First, the same message can be expressed through various emojis, each carrying subtle differences in nuance or connotation~\cite{wijeratne2017semantics}. Relying on exact match overlooks this flexibility in how meaning is conveyed through different emoji choices.
Second, this evaluation criterion may not accurately reflect the recommendation quality. Existing methods often embed individual user characteristics into the model, as recommendations are influenced by personal emoji usage patterns. Users from different demographic groups frequently use distinct emojis to convey the same sentiment or context~\cite{jones2020sex, chen2024individual, wirza2020difference}.
However, a model's ability to recommend the exact ``ground truth'' emoji may not fully reflect its understanding of the semantic richness and diversity in emoji use. Fig.~\ref{fig:semantic_preservation} provides intuitive examples of emoji recommendations. Under the traditional exact match criterion, a model would be considered incorrect if it recommends \includegraphics[height=1em]{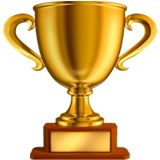} instead of \includegraphics[height=1em]{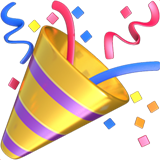} for ``I just finished a huge project!" or \includegraphics[height=1em]{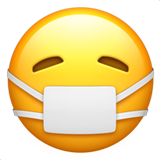} instead of \includegraphics[height=1em]{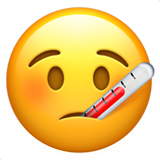} for ``Feeling a bit under the weather today." However, both \includegraphics[height=1em]{emojis_png/trophy.png}\includegraphics[height=1em]{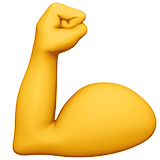}\includegraphics[height=1em]{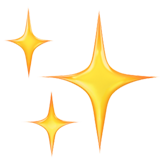} and \includegraphics[height=1em]{emojis_png/mask.png}\includegraphics[height=1em]{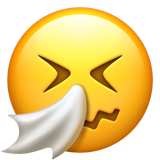}\includegraphics[height=1em]{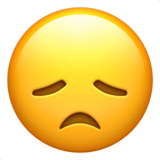}, should be treated as correct since they convey semantically consistent meanings compared with original emojis.

Therefore, instead of exact match, a more effective evaluation criterion is required---one that goes beyond merely match the user's chosen emojis and instead captures a model's capability to recommend emojis that align with the meaning of the original text.

To address these issues, we propose a novel evaluation framework for emoji recommendation: \textit{semantics preservation}. This framework aims to assess how effectively models can recommend emojis that maintain semantic consistency, ensuring that the predicted user's affective state, demographic profile, and attitudinal stance are not altered with recommended emojis. To evaluate this, we leverage five downstream classification tasks: sentiment analysis, emotion classification, stance detection, age prediction, and gender prediction.\footnote{We choose not to include racial classification as classifying races based on social media posts or emoji usage could lead to unintended bias or reinforce harmful stereotypes, which we aim to avoid.} 
The closer the classification labels of the posts with recommended emojis match those of the posts with the original emojis across these five tasks, the stronger the model's ability to perform semantics preserving emoji recommendation.

With the advent of GPT-series models~\cite{ouyang2022training}, there has been a growing trend in using LLMs to explore complex questions related to social networks~\cite{ziems2023can, zhu2023can} and digital communication behaviors, such as understanding and using emojis~\cite{lyu2024human}. LLMs are particularly well-suited for emoji recommendation, as they are not restricted to a fixed set of emojis and can generate diverse, context-appropriate recommendations for the same input. However, few studies have specifically examined LLMs' ability to perform emoji recommendation. This paper seeks to address these gaps by systematically and comprehensively evaluating the performance of LLMs, including both proprietary and open-source LLMs on the proposed semantics preserving emoji recommendation task.

The contributions of this paper are threefold:

\begin{enumerate}
    \item We introduce a novel \textit{semantics preserving} evaluation framework for emoji recommendation, specifically designed to assess the ability of models in maintaining the semantic consistency in emoji recommendations.
    
    \item We develop a comprehensive benchmark and design task-specific metrics that evaluate the semantics preserving performance of various LLMs on emoji recommendation.
    
    \item We propose several advanced prompting techniques to improve the performance of the baseline LLMs. We find that by conditioning on user profile information, the semantic consistency of LLMs' emoji recommendation can be significantly improved.
\end{enumerate}

\begin{table}[h]
\centering
\caption{Comparison between previous works and our proposed semantics preserving evaluation framework for emoji recommendation. 
\textbf{Multi-Emoji}: \lowercase{ability to recommend multiple emojis for each sentence.} 
\textbf{Open Vocab.}: \lowercase{whether the model can recommend emojis from an unrestricted set (vocabulary).} 
\textbf{Diversity}: \lowercase{ability to recommend different emojis for the same input across multiple attempts.} 
\textbf{Criterion}: \lowercase{the evaluation method used to assess emoji recommendation performance.}}
\label{tab:comparison}
\adjustbox{max width=\linewidth}{
\begin{tabular}{l|ccc|c}
\toprule
\multirow{2}{*}{\textbf{Model}} & \multicolumn{3}{c|}{\textbf{Property}}           & \multirow{2}{*}{\textbf{Criterion}} \\\cline{2-4}
 & \textbf{Multi-Emoji} & \textbf{Open Vocab.} & \textbf{Diversity} & \\ 
\midrule
DeepMoji~\cite{felbo2017using}     & \ding{55} & \ding{55} & \ding{55} & Exact match \\ 
Mojitalk~\cite{zhou2017mojitalk}      & \ding{55} & \ding{55} & \ding{55} & Exact match \\ 
mmGRU~\cite{zhao2018analyzing}        & \ding{55} & \ding{55} & \ding{55} & Exact match \\ 
Emoji Decoded~\cite{zhou2024emojis}         & \ding{55} & \ding{55} & \ding{55} & Exact match \\
Seq2Emoji~\cite{peng2021seq2emoji}     & \ding{51} & \ding{55} & \ding{55} & Exact match \\
CAPER~\cite{zhao2020caper}         & \ding{51} & \ding{55} & \ding{55} & Exact match \\
EmojiLM~\cite{peng2023emojilm}         & \ding{51} & \ding{51} & \ding{55} & Exact match \\
MultiEmo~\cite{lee2022multiemo}         & \ding{55} & \ding{55} & \ding{55} &  Category match \\
Human vs. LMMs~\cite{lyu2024human}   & \ding{51} & \ding{51} & \ding{51} & Distribution Similarity \\
\midrule
\textbf{Our work} & \textbf{\ding{51}} & \textbf{\ding{51}} & \textbf{\ding{51}} & Semantics Preservation \\ 
\bottomrule
\end{tabular}
}
\end{table}

\section{Related Work}

\subsection{Emoji Recommendation Using Traditional Methods} 
Emoji recommendation is an active research area in computational linguistics, focusing on developing models to predict emojis based on text input. Various machine learning models have been explored for these tasks. Wu~\etal~\cite{wu2018tweet} developed a model combining CNN and LSTM architectures to recommend emojis in tweets, focusing on 30 different emojis. Felbo~\etal~\cite{felbo2017using} introduced DeepMoji, a deep neural network leveraging bi-LSTM models with attention mechanisms, trained on a massive dataset of 1.2 billion tweets containing 64 common emojis. Zhou and Wang~\cite{zhou2017mojitalk} proposed a conditional variational autoencoder (CVAE) that enhances the traditional seq2seq model by incorporating additional encoders and networks, showing better performance on a dataset of 650,000 tweets. Research by Xie~\etal~\cite{xie2016neural} indicated that incorporating conversational context could improve emoji prediction accuracy, particularly in dialogues involving two individuals. Kim~\etal~\cite{kim2020no} further advanced this approach by combining machine learning with clustering to recommend contextually appropriate emojis based on conversation topics and emotions. Lee~\etal~\cite{lee2022multiemo} proposed MultiEmo, a multi-task framework that considers emotion detection to predict the most relevant emoji. They also introduced new evaluation metrics for emoji category prediction, showing superior performance on the Twitter dataset compared to existing models.

Beyond text-based modeling, some research has increasingly explored the use of multiple modalities. These methods combine images, demographics, time, and location, in conjunction with textual data to predict emojis. Barbieri~\etal~\cite{barbieri2018multimodal} employed a multimodal approach using pictures, text, and emojis in Instagram posts, developing models based on bi-LSTM for text and ResNet for visual data, demonstrating that a combination of modalities outperforms text-only models. Zhao~\etal~\cite{zhao2018analyzing} proposed a multitask, multimodal gated recurrent unit (mmGRU) model that incorporates text, images, and user demographics to predict emoji categories and their positions within a text, revealing regional differences in emoji usage and understanding. Additional studies, such as those by Barbieri~\etal~\cite{barbieri2018exploring}, explored the temporal aspects of emoji usage, training models on seasonal data to observe changes in emoji semantics. Zhao~\etal~\cite{zhao2020caper} also utilized contextual information, including user preferences, user gender, current time and emoji co-occurrence factors, to refine emoji recommendations across platforms like Sina Weibo and Twitter (now renamed to X). These studies primarily emphasize exact emoji matching, which, as previously mentioned, has its limitations. In contrast, our evaluation framework focuses on preserving semantic consistency.

\subsection{Emoji Recommendation Using Large Language Models}
With the advent of large language models (LLMs), many studies have begun to explore using LLMs for emoji-related research. Kheiri~\etal~\cite{kheiri2023sentimentgpt} examined various GPT-based strategies, such as prompt engineering, fine-tuning, and embedding classification, for sentiment analysis, demonstrating a significant improvement in predictive performance—over 22\% in F1-score compared to state-of-the-art models. Zhou~\etal~\cite{zhou2024emojis} assessed ChatGPT's effectiveness in emoji-related tasks, suggesting it could serve as an alternative to human annotators by providing clear explanations and reducing misunderstandings in online communication. Lyu~\etal~\cite{lyu2024human} investigated GPT-4V's interpretation and usage of emojis, finding that it exhibits less interpretative ambiguity and a broader variety in emoji selection compared to humans, indicating partial alignment with human usage patterns. Peng~\etal~\cite{peng2023emojilm} built a parallel corpus for text-to-emoji translation using GPT-3.5 Turbo and develops a specialized sequence-to-sequence model for bidirectional translation between text and emojis. However, these studies primarily focus on the capabilities of specific LLMs like GPT in emoji research. In our work, we \textit{extend this exploration to multiple LLMs} for emoji recommendation and propose novel metrics to evaluate their performance more comprehensively.

\subsection{Comparison with Prior Work}

Compared to previous works, our semantics preserving emoji recommendation task introduces an innovative criterion designed to assess how effectively models recommend emojis that maintain \textbf{semantic consistency}. Additionally, our framework permits three key properties of emoji recommendation methods, as outlined in Table~\ref{tab:comparison}.

First, unlike earlier works that typically recommend only a single emoji per sentence, our new task allows \textbf{multi-emoji} recommendation, which more accurately mirrors real-world usage where users often combine multiple emojis to express nuanced sentiments. Second, prior approaches often restrict the emoji vocabulary to a predefined set of commonly used emojis. For instance, mmGRU limits recommendations to 35 frequently used emojis per tweet, and CAPER selects from a list of the 50 most popular emojis. In contrast, our task supports an \textbf{open emoji vocabulary}, allowing models to recommend any emoji relevant to the text. 
Finally, our task allows emoji recommendation methods to suggest \textbf{diverse}, contextually appropriate emojis, reflecting a broader range of expressions that better align with real-world usage.

\begin{figure*}[t]
    \centering
    \includegraphics[width=\textwidth]{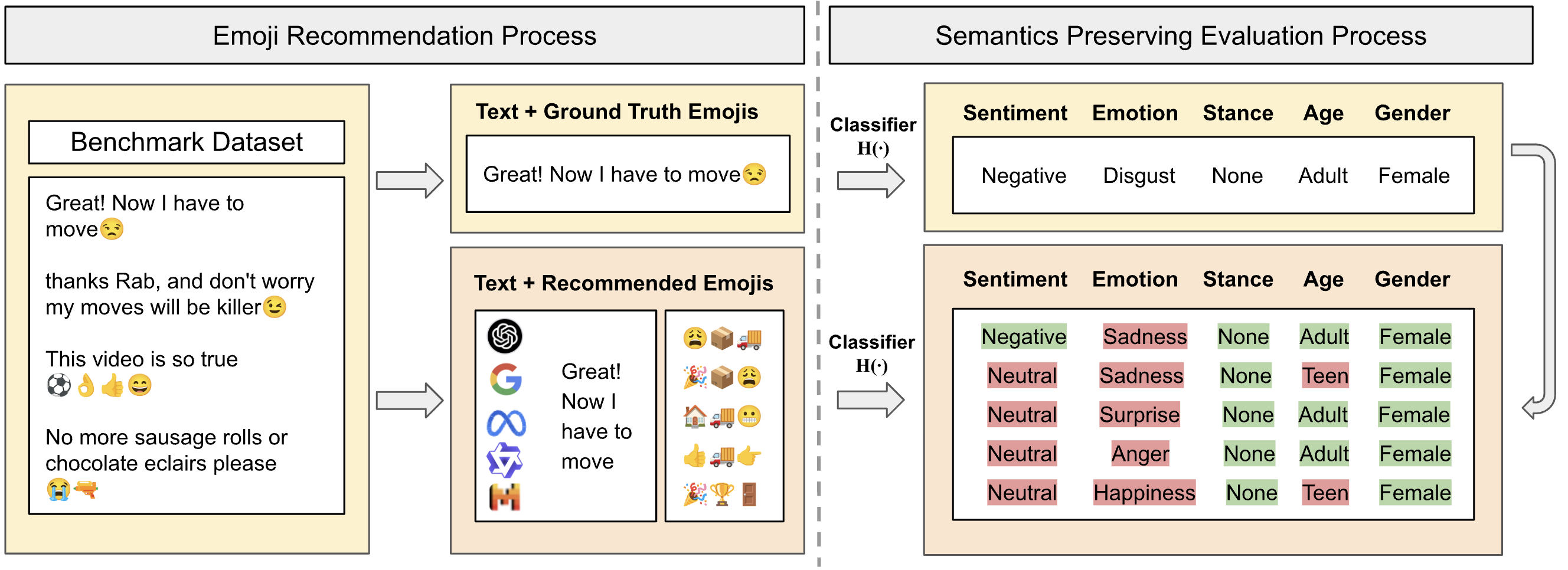} 
    \caption{Overview of the Semantics Preserving Emoji Recommendation Framework. Left side: The Emoji Recommendation Process uses large language models to recommend three emojis for texts from the benchmark dataset. Right side: The Semantics Preserving Evaluation Process compares text + predicted emojis with text + ground truth emojis across 5 selected semantic dimensions, including sentiment, emotion, stance, age, and gender.
}
    \label{fig:emoji_recommendation_framework}
\end{figure*}

\section{Semantics Preserving Emoji Recommendation}
\label{sec: framework}

\textbf{Task Formulation:}
Given the plain text of the original user post denoted as \( T \) and the original emojis associated with this text as \( E_{\text{ori}} \), a model is tasked with recommending $n$ emojis, denoted as \( E_{\text{rec}}\) that maintain the semantic content of the original text, as illustrated in the left part of Fig.~\ref{fig:emoji_recommendation_framework}.

\textbf{Recommendation Evaluation:}
 For each sentence with the predicted or ground truth emojis, we use a strong pre-trained classifier $H(\cdot)$ to infer labels on five downstream tasks. We assess whether the inferred affective state, demographic profile, and attitudinal stance of the user remain unchanged as shown in the right part of Fig.~\ref{fig:emoji_recommendation_framework}. If these attributes are preserved, we consider the recommended emojis to have maintained the original semantics. 

 Specifically, we first use the classifier $H(\cdot)$ to assign labels to the original sentence combining \( T \) with \( E_{\text{ori}} \) on five downstream tasks: sentiment analysis, emotion classification, stance detection, age prediction, and gender prediction. The classification output for each downstream task \( d \) (where \( d \in \{ \text{sentiment, emotion, stance, age, gender} \} \)) is denoted as \( H_{d}(T, E_{\text{ori}}) \), which serves as the ground truth. Next, we use the same classifier to assign labels to the modified sentences that combine the original plain text \( T \) with the recommended emojis \( E_{\text{rec}}\). The classification output for the sentence with the recommended emojis is represented as \( H_{d}(T, E_{\text{rec}}) \). For each downstream task \( d \), the semantics preserving capability of the model is evaluated by comparing whether \( H_{d}(T, E_{\text{rec}}) \) equals the ground truth \( H_{d}(T, E_{\text{ori}}) \). 

\subsection{Selection of Downstream Tasks}
A critical aspect of evaluating semantics preservation in emoji recommendation is the selection of appropriate downstream tasks on social media~\cite{lyu2023gpt}. We propose that, for a recommended emoji to achieve semantics preservation, it should maintain two key types of information consistent with the original emojis: emotional content and user demographics.

The first type of information is the emotional content conveyed by the sentence. To capture this aspect, we select three downstream tasks: sentiment analysis, emotion classification, and stance detection. Emojis often serve as indicators of \textbf{sentiment analysis}~\cite{shiha2017effects,chen2020image}, so maintaining the sentiment label after replacing the original emoji with a recommended one is crucial for semantics preservation. For this task, we use sentiment labels such as positive, negative, and neutral to evaluate consistency. \textbf{Emotion classification} goes beyond general sentiment by identifying specific emotions expressed in the text. Following Ekman's six basic emotions~\cite{ekman1992there}, we classify emotions into anger, disgust, fear, joy, sadness, and surprise. We expect that a recommended emoji should align with the six emotional contexts of the original sentence. \textbf{Stance detection} is about identifying the author's position or attitude towards a topic. Emojis can modify or reinforce the stance expressed in a sentence, so it is essential that the recommended emojis preserve the stance conveyed by the original text~\cite{zotova2021semi}. We classify stance using the labels none, favor, and against.

The second type of information relates to the user's demographic characteristics. For this part, we select two downstream tasks: age prediction and gender prediction. We opt not to include racial classification as a downstream task to keep the focus of our benchmark on broader aspects of semantics preservation, without introducing sensitive demographic factors. For \textbf{age prediction} task, rather than specific ages, we categorize users into broader groups: child, teen, adult, and senior~\cite{nguyen2014gender}. Different age groups have distinct patterns in emoji usage, even when conveying the same semantic content.~\cite{jaeger2018emoji} Ensuring that the recommended emoji maintains the same age classification as the original helps confirm semantics preservation. For \textbf{gender prediction} task, research indicates notable differences in how different genders use emojis to convey the same meaning~\cite{koch2022age}. The gender labels are classified as male and female. We recognize that gender identity exists on a spectrum, including non-binary, and other gender non-conforming identities, which are frequently underrepresented in existing datasets. However, to maintain the focus on the core objectives of semantics preservation and emoji recommendation, we limit gender prediction to two categories in this study. We encourage the research community to develop and use datasets that reflect the full diversity of gender identities to help mitigate the perpetuation of existing biases. 

By evaluating how the predicted emojis align with the original emojis across these five downstream tasks, we can comprehensively evaluate whether the recommended emojis achieve semantics preservation.

\subsection{Downstream Task Classifier}

To effectively assess semantics preservation via measuring the performance on downstream tasks, selecting a strong classifier for downstream tasks is essential. Large Language Models have recently emerged as powerful tools, often matching or surpassing human-level classification accuracy across various tasks~\cite{zheng2023judging,lyu2023gpt}. In this section, we compare the performance of GPT as a classifier against the traditional top-performant BERTweet model~\cite{nguyen2020bertweet} in classification tasks. In each experiment, to mitigate randomness, we query GPT-4o-mini three times and use the majority label as the final prediction. For each query, we set the temperature parameter to 0, while all other hyperparameters as their default values.

\subsubsection{Accuracy of GPT as a Classifier}

We conduct a preliminary experiment focusing on one of the downstream tasks: gender classification. For this experiment, we extract 2,000 tweets with authentic gender labels from the PAN18 dataset~\cite{rangel2018overview} for training set and 1,000 tweets for the testing set. BERTweet is fine-tuned on the training set to function as a gender classifier. For the GPT models, both GPT-4o and GPT-4o-mini are prompted to classify gender using the following instruction: \textit{``I will provide you a tweet. Please classify the likely gender of the person who wrote the tweet as 'male' or 'female'. Please only output the answer without justification. Here is the tweet: {text}''}

Table~\ref{tab:gender_classification_performance} presents the gender classification performance of the three models on the test set, including accuracy, precision, recall, and F1-score. The results demonstrate that both GPT-4o and GPT-4o-mini outperform BERTweet in gender classification accuracy.

\subsubsection{Sensitivity of GPT to emojis}

While GPT-4o and GPT-4o-mini achieve a higher accuracy than BERTweet in gender classification, it is important to assess their sensitivity to emojis before adopting GPTs as the classifier for our downstream tasks. To explore this, we design another experiment and assess how emojis affect the GPT models' performance.

In this experiment, we take the original set of tweets used in the gender classification task and replace the existing emojis with three random emojis, selected from the top 500 frequently used emojis from Unicode\footnote{Available at: \url{https://home.unicode.org/}}. We then prompt GPT-4o and GPT-4o-mini to predict the gender labels for both the original tweets and the tweets with random emojis, using the same instruction format as the first experiment.

To evaluate the consistency of the models' predictions between the two sets of tweets, we calculate the Cohen's Kappa value and correlation between the two predicted label sets. Table~\ref{tab:GPT_sensitivity} shows that the Kappa value and correlation of GPT-4o-mini is lower than that of GPT-4o, indicating that the predictions of GPT-4o-mini are more affected by the introduction of random emojis. This suggests that GPT-4o-mini is more sensitive to emojis compare to GPT-4o.

In addition to this experimental result, several studies have demonstrated the effectiveness of GPT models as classifiers for sentiment, emotion, stance, age, and gender. For instance, Rathje~\etal~\cite{rathje2024gpt} showed that GPT models achieved high accuracy in detecting psychological constructs, such as sentiment and discrete emotions, across multiple languages, often outperforming traditional dictionary-based methods and rivaling top-performing fine-tuned models. Similarly, Liyanage~\etal~\cite{liyanage2023gpt} evaluated GPT-4 as a stance classifier and found that it performs comparably to human annotators.

Based on the experimental findings and supporting evidence from existing literature, we select GPT-4o-mini as the classifier for our downstream tasks due to its strong performance and cost-effectiveness.

\begin{table}[h]
\centering
\caption{Gender classification performance of BERTweet, GPT-4o and GPT-4o-mini. Best results are in bold.}
\label{tab:gender_classification_performance}
\begin{tabular}{l|ccc}
\toprule
\textbf{Model}   & \textbf{Precision} & \textbf{Recall} & \textbf{F1-Score} \\
\midrule
BERTweet         & 0.679              & 0.620           & 0.633             \\ 
GPT-4o-mini      & 0.728              & 0.737           & 0.701             \\ 
GPT-4o           & {\bf 0.758}              & {\bf 0.764}           & {\bf 0.741}             \\ 
\bottomrule
\end{tabular}
\end{table}

\begin{table}[h]
\centering
\caption{Sensitivity of different GPT versions to emojis, where lower values indicate higher sensitivity to emojis.}
\label{tab:GPT_sensitivity}
\begin{tabular}{l|cc}
\toprule
\textbf{Model}   & \textbf{Cohen's Kappa} & \textbf{Correlation} \\
\midrule
GPT-4o-mini      & {\bf 0.470}                  & {\bf 0.474}                \\ 
GPT-4o           &  0.626                 &  0.627                \\ 
\bottomrule
\end{tabular}
\end{table}

\subsection{Downstream Task-based vs. Human Judgement-based}
After validating the effectiveness of GPT-4o-mini as a classifier, another key aspect that needs to be examined is the effectiveness of the proposed downstream task-based semantics preserving evaluation framework. Accordingly, we compare the semantics preservation scores based on the five downstream tasks with human judgement.

We begin by calculating the number of matching pairs between the original sentence (text + ground truth emojis), and the sentence with emojis generated by GPT-4o (text + recommended emojis). The matching pairs represent the number of labels that are the same across the five downstream tasks, with value ranging from 0 to 5. We use the proportion of correctly matched labels as the downstream task-based semantics preservation score, as shown in Table~\ref{tab:human_annotation}. To ensure a balanced evaluation, we select all data where the number of matching pairs are 0 or 1, and randomly select 50 samples from the data where the number of matching pairs is 2, 3, 4 or 5.

The evaluation dataset is then independently annotated by four researchers. Each researcher performs a binary classification, where a predicted sentence is labeled as 1 if it preserves semantic meaning of the original sentence , and 0 otherwise. The average score for each data sample is then calculated to represent human judgement. 

The results, as detailed in Table~\ref{tab:human_annotation}, show that the trends in both the downstream task-based semantics preservation scores and human judgement-based scores are correlated. This alignment further validates our method in evaluating semantics preserving emoji recommendation using downstream tasks.

\begin{table}[h]
\centering
\caption{Comparison between the downstream task-based and human judgement-based semantics preservation scores.}
\label{tab:human_annotation}
\adjustbox{max width=\linewidth}{
\begin{tabular}{c|ccc}
\toprule
\textbf{Matching Pairs} & \textbf{Count} & \textbf{Downstream task-based} & \textbf{Human judgement-based} \\ 
\midrule
0   & 4   & 0.0   & 0.25  \\ 
1   & 65   & 0.2   & 0.442 \\ 
2   & 50   & 0.4   & 0.46  \\
3   & 50   & 0.6   & 0.63  \\ 
4   & 50  & 0.8   & 0.745 \\ 
5   & 50   & 1.0   & 0.85  \\ 
\bottomrule
\end{tabular}}
\end{table}

\section{Benchmark Construction Pipeline}

We construct our benchmark dataset based on the PAN18 dataset~\cite{rangel2018overview}. Our goal is to create a high quality dataset that allows us to evaluate the emoji recommendation performance of LLMs on various downstream emoji-related text classification tasks, including sentiment analysis, emotion classification, stance detection, age prediction, and gender prediction.

\subsection{Dataset Filtering}

We begin by filtering the English subset of the PAN18 dataset to include only tweets containing one of the top 500 most frequently used emojis, as listed on the Unicode Website\footnotemark[2]. Additionally, duplicate entries are removed, maintaining a diverse set of 20,827 tweets for analysis.

\subsection{Labeling Data for Downstream Tasks}

To effectively test the model's ability to recommend emojis across different semantic contexts, we aim for a balanced distribution of labels in the dataset. Therefore, we first employ GPT-4o-mini to classify each of the 20,827 tweets on each of the five downstream tasks. After obtaining the initial labels, we then remove instances where the GPT-4o-mini's predicted label does not align with the predefined class names of the downstream task, ensuring the reliability of the dataset labels. 

\subsection{Dataset Balancing}
The distribution of labels after this filtering process is shown in Fig.~\ref{fig:final_distribution}. It is evident that the label distribution in each task is highly imbalanced, particularly in tasks where the number of classes exceeds two, such as emotion classification, age prediction, and stance detection.
To address the label imbalance issue, we apply a down-sampling strategy on the filtered data. Specifically, we extract all instances of the minority classes, including: 
\begin{enumerate}

\item All tweets labeled as \textit{disgust}, \textit{anger}, and \textit{fear} for the emotion classification task.
    \item All tweets labeled as \textit{child} and \textit{senior} for the age prediction task.
    \item All tweets labeled as \textit{against} for the stance detection task.
\end{enumerate}

Additionally, we randomly select 2,000 instances from the \textit{male} gender class to further balance the gender distribution. After applying these sampling techniques, we compile a benchmark dataset comprising 5,039 tweet posts with an average of 2.25 emojis per sentence. The average sentence length is 14.2 tokens. The final distribution of data across the five tasks is shown in Fig.~\ref{fig:final_distribution}. The resulting benchmark dataset provides a set of examples that are more balanced across all the downstream tasks, enabling a comprehensive evaluation of the emoji recommendation models' ability to maintain semantic integrity across diverse contexts and user demographics.

\begin{figure}
    \centering
    \includegraphics[width=\linewidth]{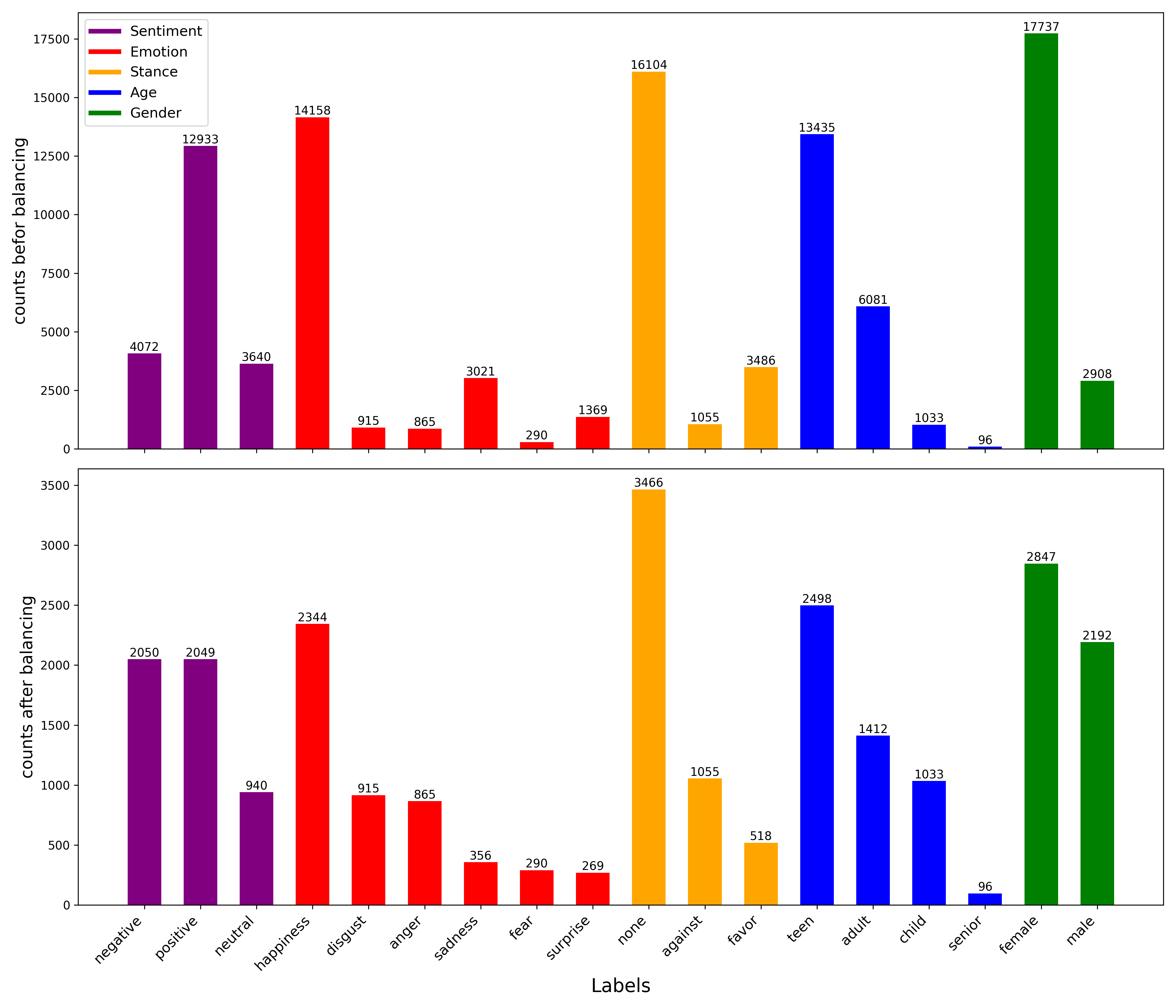}
    \caption{Comparison of data distribution across five downstream tasks in the benchmark dataset before and after balancing.}
    \label{fig:final_distribution}
\end{figure}

\section{Experiments}
In this section, we evaluate the performance of existing large language models for the semantics preserving emoji recommendation task on our constructed benchmark dataset. We also provide a detailed analysis on the models' performance across the five downstream tasks and assess the diversity of the emojis used by the models.

\subsection{LLMs as Emoji Recommendation Models}

We select the state-of-the-art LLM models as the emoji recommendation methods, focusing on both open-source and proprietary options. For open-source models, we choose LLaMa3.1-70B-Instruct\footnote{Available at: \url{https://www.llama.com/}}, LLaMa3.1-8B-Instruct\footnotemark[4], Qwen2-72B-Instruct~\cite{yang2024qwen2}, Gemma2-9B-Instruct~\cite{team2024gemma}, and Mistral-7B-Instruct~\cite{jiang2023mistral} with parameter sizes ranging from 7 billion to 70 billion, due to their proven performance in various natural language processing tasks. For proprietary models, we selected GPT-4o~\cite{ouyang2022training}, the most recent model at the time of writing from OpenAI, known for its exceptional performance across various benchmarks. By utilizing both open-source and proprietary models across a range of parameter size, we ensure a comprehensive evaluation of the emoji recommendation performance. 

We also randomly assign three emojis to each sentence as a baseline to compare the performance of model-generated recommendations against random selection in preserving the semantics of the text.

\begin{table*}[ht]
\centering
\caption{Precision, Recall and F1 Scores (all in \%) and Downstream Tasks' Accuracy comparison of various LLMs. Avg denotes to the average precision across all five tasks, \ie the semantics preserving score. Best results are in bold.}
\label{tab:main_cls_results}
\begin{tabular}{|l|c|c|c|c|c|c|c|c|c|}
\hline
\textbf{LLM} & \textbf{Precision} & \textbf{Recall} & \textbf{F1 score} & \textbf{Sentiment} & \textbf{Emotion} & \textbf{Stance} & \textbf{Age} & \textbf{Gender} & \textbf{Avg} \\
\hline

\multicolumn{10}{|c|}{\textbf{Zero-shot}} \\
\hline
GPT-4o & 6.93   & 16.14 & 9.04 & 78.41 & \textbf{67.41} & \textbf{88.61} & \textbf{82.02} & 79.70 & \textbf{79.23}  \\

LLaMa3.1-70B-Instruct & 7.26   & \textbf{16.89} & \textbf{9.52} & 78.27 & 67.18 & 87.24 & 81.76 & 80.39 & 78.97  \\

Gemma2-9B-Instruct & 6.71   & 15.61 & 8.70 & \textbf{78.57} & 66.44 & 87.42 & 80.59 & \textbf{81.78} & 78.96  \\

Qwen2-72B-Instruct & \textbf{7.30}   & 14.99 & 9.16 & 76.80 & 63.41 & 87.78 & 81.52 & 79.86 & 77.87  \\

LLaMa3.1-8B-Instruct & 5.69   & 13.17 & 7.44 & 76.32 & 65.71 & 85.67 & 78.11 & 80.49 & 77.26  \\

Mistral-7B-Instruct & 3.70   & 7.37 & 4.56 & 76.68 & 60.83 & 86.45 & 78.43 & 75.90 & 75.66  \\

Random Sentence & 0.27   & 0.52 & 0.33 & 69.12 & 57.09 & 86.96 & 75.17 & 74.88 & 72.65  \\

\hline

\multicolumn{10}{|c|}{\textbf{Few-shot}} \\
\hline
GPT-4o & 7.39  & 17.67  & 9.70 & \textbf{78.79}  & \textbf{67.85} & \textbf{88.13}  & 81.68  & 77.95 & 78.88  \\ 
LLaMa3.1-70B-Instruct & \textbf{8.17} & \textbf{19.50} & \textbf{10.86} & 78.45  & 67.43  & 86.25  & 82.54  & 80.43  &  79.02  \\

Gemma2-9B-Instruct & 7.67 &  17.68 & 9.97  & 78.59  & 66.74  & 87.97  & 80.33  & \textbf{82.36} &  \textbf{79.20} \\

Qwen2-72B-Instruct & 8.14 & 16.02  & 10.04 & 77.06 & 63.62 & 87.93 & \textbf{82.56} & 81.13  & 78.46  \\

LLaMa3.1-8B-Instruct & 6.63 & 16.11  & 8.84 & 77.63 & 66.62 & 87.16 & 80.99 & 79.96  &  78.47 \\

Mistral-7B-Instruct & 4.21 & 9.08  & 5.30  & 77.40 & 61.94 & 88.01 & 80.67 & 77.22  &  77.05 \\
\hline

\multicolumn{10}{|c|}{\textbf{Conditional Recommendation}} \\
\hline
GPT-4o & 8.23 & 19.5  & 10.8 & 78.61  & 67.49 & 88.33  & 85.69  & 87.74 & 81.57  \\

LLaMa3.1-70B-Instruct  & \textbf{8.56} & \textbf{20.27} & \textbf{11.34} & 78.41  & \textbf{68.01}  & 87.70  & 85.04  & 87.61  & 81.35 \\

Gemma2-9B-Instruct & 8.31 & 19.9  & 10.96 & \textbf{79.48}   &  67.73 & 87.72  & \textbf{86.58}  & \textbf{87.89}  & \textbf{81.88}  \\

Qwen2-72B-Instruct & 8.54 & 17.11  & 10.58  & 75.95 & 62.43 & 87.64 & 84.54 & 87.76 &  79.66 \\

LLaMa3.1-8B-Instruct & 7.08 & 16.39  & 9.29  & 77.16 & 66.82 & 86.33 & 83.39 & 87.82  &  80.30 \\

Mistral-7B-Instruct & 3.60 & 8.72  & 4.69  & 78.29 & 62.69 & \textbf{88.37} & 82.73 & 84.24  &  79.26 \\
\hline

\end{tabular}
\end{table*}

\subsection{Prompting strategies of LLMs for Emoji Recommendation}

When using large language models to predict emojis, prompt design plays a critical role in ensuring accurate and contextually relevant results. In this study, we explor three different prompting strategies: 1) Zero-shot, 2) Few-shot, and 3) Conditional Generation. For all strategies, we set the temperature to the default value of $1$, which encourages the model to make confident predictions with some level of randomness and creativity in the emoji recommendations. For the open-source models, we access them via the Together AI\footnote{Available at: \url{https://www.together.ai/}} platform and make API calls to generate predictions.

\textbf{Zero-shot Prompting} In the zero-shot approach, the model is provided with a task-specific instruction without any examples. The prompt contains only instructions explaining what the model should do, and the input is a sentence without any predefined labels or additional guidance. For each user \textit{text}, the prompt inputted to the LLMs is as follows: \textit{"As a social media user, find three emojis that best fit the context: \{text\}. Do not include any words or explanations—only three emojis."}

\textbf{Few-shot Prompting} Few-shot prompting builds upon the zero-shot approach by providing the model with a few examples before asking it to predict emojis. Each example will demonstrate how to pair text with emojis to help the LLM model better understand the task.

Specifically, we employ a 5-shot approach, carefully selecting diverse examples to cover a wide range of categories across the five tasks. In these five examples, we include texts and emojis representing various emotions, age groups, and genders to ensure each task could learn from this diversity. These examples are sourced from real users and are not part of the benchmark dataset. The five examples are as follows: 

\begin{enumerate}
\item Context: "After the success of  's walima vid, we're back at it again for  's wedding celebrations!"
   Emojis: \includegraphics[height=1em]{emojis_png/tada.png}\includegraphics[height=1em]{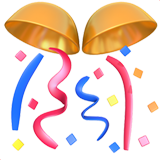}\includegraphics[height=1em]{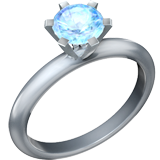}

\item Context: ". finally being played in the right position as well. Great performance!"
   Emojis: \includegraphics[height=1em]{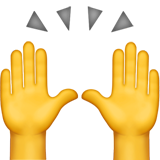}\includegraphics[height=1em]{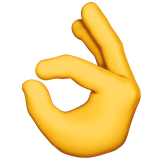}\includegraphics[height=1em]{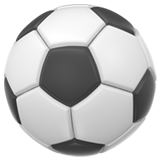}

\item Context: "I NEED THE WILSON BLADE 104 IN MY LIFE!!!"
   Emojis: \includegraphics[height=1em]{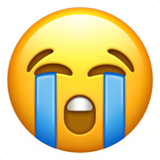}\includegraphics[height=1em]{emojis_png/sob.png}\includegraphics[height=1em]{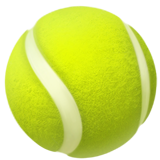}

\item Context: "Just broke me tooth into bits eating a lid of a bottle"
   Emojis:\includegraphics[height=1em]{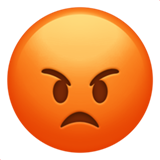}\includegraphics[height=1em]{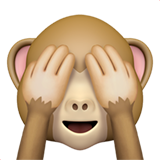}\includegraphics[height=1em]{emojis_png/rage.png}

\item Context: "It's not even 9am and it's already 30 degrees"
   Emojis:\includegraphics[height=1em]{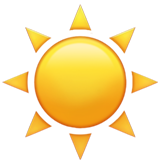}\includegraphics[height=1em]{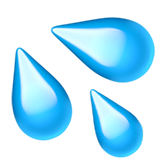}\includegraphics[height=1em]{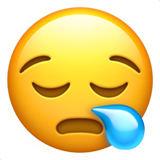}
\end{enumerate}

\textbf{Conditional Emoji Recommendation}
In the previous zero-shot or few-shot prompting approaches, the input prompt to the LLM is either the original tweet post, or adding predefined examples. These approaches ignore the rich information that could be obtained from the user's profile under certain circumstances. To further improve the recommendation performance, we propose conditional emoji recommendation, where we embed user's gender and age labels as additional context into the prompt to guide the prediction of the emojis. The gender and age labels are predicted by GPT from the original tweet post as a simulation of the real world scenario, where we have access to the demographic information from the user profile. We evaluate whether this conditional recommendation approach can lead to more semantically consistent emojis predictions. 

\begin{table*}[htbp]
\centering
\caption{Correlation Between Emoji Recommendation and Semantics Preservation Accuracy: A Comparative Analysis of GPT-4o and Mistral Models in Zero-Shot Settings.  Red cells indicate mismatches, while green cells indicate matches with the ground truth.}
\label{tab:emoji_compare}
\resizebox{\textwidth}{!}{
\begin{tabular}{|l|c|c c c c c|}
\hline
\textbf{Model} & \textbf{Sentence with emojis} & \multicolumn{5}{|c|}{\textbf{Labels on five tasks}}   \\ 
\hline

Ground Truth & Roll on the prem at the weekend, international football is dire 
\includegraphics[height=1em]{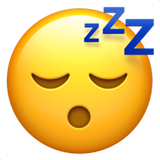}
 & Negative & Disgust &  Against & Adult    & Male \\

GPT-4o &  Roll on the prem at the weekend, international football is dire \includegraphics[height=1em]{emojis_png/soccer.png} \includegraphics[height=1em]{emojis_png/sleeping.png} \includegraphics[height=1em]{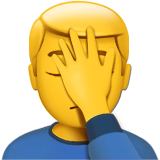}  & \colorbox{green!12}{Negative} & \colorbox{red!10}{Sadness} & \colorbox{green!12}{Against}  & \colorbox{green!12}{Adult}   & \colorbox{green!12}{Male}  \\

Mistral-7B-Instruct &  Roll on the prem at the weekend, international football is dire \includegraphics[height=1em]{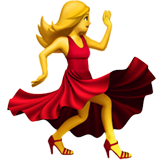} \includegraphics[height=1em]{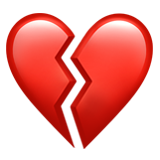} \includegraphics[height=1em]{emojis_png/trophy.png}  & \colorbox{green!12}{Negative} & \colorbox{red!10}{Sadness}  &  \colorbox{red!10}{None} & \colorbox{red!10}{Teen} & \colorbox{red!10}{Female}\\

\hline
Ground Truth & C'mon Dodgers I need to see you guys in the World Series \includegraphics[height=1em]{emojis_png/muscle.png}
\includegraphics[height=1em]{emojis_png/muscle.png}
\includegraphics[height=1em]{emojis_png/muscle.png} & Positive & Happiness &  Favor & Teen  & Male   \\

GPT-4o &  C'mon Dodgers I need to see you guys in the World Series \includegraphics[height=1em]{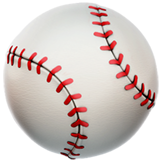}
\includegraphics[height=1em]{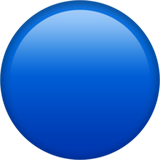}
\includegraphics[height=1em]{emojis_png/trophy.png}  & \colorbox{green!12}{Positive} & \colorbox{green!12}{Happiness}  &  \colorbox{green!12}{Favor}  & \colorbox{green!12}{Teen} & \colorbox{green!12}{Male}  \\

Mistral-7B-Instruct &  C'mon Dodgers I need to see you guys in the World Series \includegraphics[height=1em]{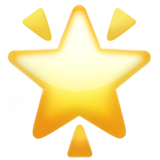}
\includegraphics[height=1em]{emojis_png/trophy.png}
\includegraphics[height=1em]{emojis_png/broken_heart.png} & \colorbox{red!10}{Neutral} & \colorbox{red!10}{Sadness} &  \colorbox{green!12}{Favor} & \colorbox{green!12}{Teen}  & \colorbox{red!10}{Female}   \\

\hline
Ground Truth & you should really keep your spurts to yourself 
 \includegraphics[height=1em]{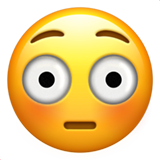}  & Negative & Disgust &  None & Teen   & Female \\

GPT-4o &  you should really keep your spurts to yourself 
 \includegraphics[height=1em]{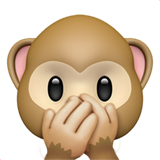}
\includegraphics[height=1em]{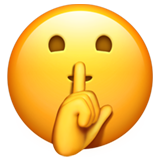}
\includegraphics[height=1em]{emojis_png/flushed.png}  & \colorbox{green!10}{Negative} & \colorbox{green!10}{Disgust} &  \colorbox{green!12}{None}  & \colorbox{green!12}{Teen} & \colorbox{green!10}{Female} \\

Mistral-7B-Instruct &  you should really keep your spurts to yourself  \includegraphics[height=1em]{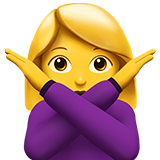}
\includegraphics[height=1em]{emojis_png/shushing_face.png}
\includegraphics[height=1em]{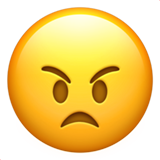} & \colorbox{green!10}{Negative} & \colorbox{red!10}{Anger}  &  \colorbox{red!12}{Against} & \colorbox{green!12}{Teen} & \colorbox{red!10}{Male}   \\

\hline
\end{tabular}
}
\end{table*}

\subsection{GPT-4o-mini as Task Classifier and Label Generator}
For evaluating semantics preservation in emoji recommendation, we choose GPT-4o-mini as the classifier for downstream tasks concerning recommendation accuracy, efficiency and cost among a set of LLM models. 

For each sentence in the dataset, we feed the text and corresponding emojis to GPT-4o-mini, prompting it to generate labels for each of the five tasks, with a temperature setting of 0. Below is an example of the prompt used for sentiment analysis: 
\textit{"I will provide you a tweet. Please classify the sentiment of the tweet as positive, negative, or neutral. Please only output the answer without justification. Here is the tweet:\{text\}"}. The same approach is adapted for emotion, stance, age, and gender classification, each with task-specific classes. These generated labels from the original post will be used as the ground truth label for each post. 

For each predicted sentence (original text + predicted emojis), we also use GPT-4o-mini to predict its labels across the five downstream tasks. Then we compare the difference between the predicted labels and ground truth label.

\subsection{Evaluation Metrics}

For each user sentence, we use metrics such as F1-score, precision, and recall as the traditional emoji prediction quantitative metrics to evaluate the performance of models in emoji recommendation. These metrics compares the extent of exact match between the model predicted emojis and the ground truth emojis in in sentence.  In our work, as described in Section~\ref{sec: framework}, we propose a novel evaluation approach, \textit{semantics preserving score}, to specifically measure the semantics preserving capabilities of the models. On each downstream task, the label of the sentence with the predicted emojis are compared with the label of the original sentence, and we calculate the proportion of correctly predicted cases as the semantics preserving score for this task. The overall semantics preserving score is obtained by averaging the scores across all tasks.

\begin{table*}[h]
\centering
\caption{Comparison of models on various downstream classification tasks using three approaches: zero-shot, few-shot, and conditional recommendation (CR). The Avg column represents the mean accuracy across five tasks. Bold values indicate the best performance for each category.}
\label{tab:detail_cls}
\resizebox{\textwidth}{!}{  
\begin{tabular}{|l|c|c|c|c|c|c|c|c|c|c|c|c|c|c|c|c|c|c|c|}
\hline
\multirow{2}{*}{\textbf{LLM}} & \multicolumn{3}{c|}{\textbf{Sentiment}} & \multicolumn{6}{c|}{\textbf{Emotion}} & \multicolumn{3}{c|}{\textbf{Stance}} & \multicolumn{4}{c|}{\textbf{Age}}  & \multicolumn{2}{c|}{\textbf{Gender}}  & \multirow{2}{*}{\textbf{Avg.(5)}} \\
\cline{2-19}
 & \textbf{Positive} & \textbf{Negative} & \textbf{Neutral}  & \textbf{Happiness} & \textbf{Sadness} & \textbf{Fear} & \textbf{Anger} & \textbf{Surprise} & \textbf{Disgust}& \textbf{Favor} & \textbf{None} & \textbf{Against} & \textbf{Child} & \textbf{Teen} & \textbf{Adult} & \textbf{Senior}   & \textbf{Male} & \textbf{Female}  & \\

\hline
GPT-4o  & 87.51 & 81.37 & 52.13  & 84.00 & 57.30 & 61.38 & 67.05 & \textbf{45.35} & 37.60 & 78.38  & 91.78 & 83.22 & 70.28 & 91.91 & 77.27 & 20.83   & 70.30   & 86.93  & 79.23\\

GPT-4o\textsubscript{few-shot}  & 88.43 & 81.85 & 51.06 & 85.71 & 59.55 & 59.31  & 67.86 & 40.52 & 36.07 & \textbf{78.76} & 91.63 & 81.23 & 66.51  & 91.63 & 79.25 & 21.88  & 68.61 & 85.14 & 78.88 \\

GPT-4o\textsubscript{CR}   & 91.17  & 80.88 & 46.28 & 87.97 & 57.30 & 53.45  & 63.82 & 33.09 & 37.05  & 77.41 & 92.04 & 81.52 & 79.38  & 94.20 & 79.24 & 27.08 & 75.00  & \textbf{97.54}  & 81.57 \\

\hline
LLaMa3.1-70B-Instruct   & 86.14  & 84.68 & 47.13  & 82.81 & 53.93 & 63.79 &  67.17 & 40.52  &  41.20 & 74.90 & 89.56 & 85.69  & 67.57  &  92.95 & 76.49 & 20.83   & 70.30 & 88.16 & 78.97\\

LLaMa3.1-70B-Instruct\textsubscript{few-shot}  &  88.92 & 83.61 & 44.36 & 85.54 & 52.25  & 57.59 & 67.05 & 32.34 & 40.77 & 75.87 & 90.25 & \textbf{86.25} & 70.18 & 92.99 & 77.34 & 19.76  & 67.75 & 90.20  & 79.20\\

LLaMa3.1-70B-Instruct\textsubscript{CR} & 89.70 & 83.46 & 42.87 & 86.90 & 52.80 & 59.65 & 65.43 & 27.51  & \textbf{42.51} & 77.99 & 89.87 & 85.31  & \textbf{80.83} & 93.95 &  76.49 & 23.96   & 75.13 & 97.22  & 81.35 \\

\hline

Gemma2-9B-Instruct  & 86.43 & 84.00 & 49.57  & 82.34 & 53.93 & 60.34 & 71.21 & 42.01 & 35.19 & 76.83 & 90.19 & 83.51   & 66.31 & 91.67 & 75.42 & 21.87  & 67.70 & 92.63  & 78.96\\

Gemma2-9B-Instruct\textsubscript{few-shot}   & 86.97 & 83.80 & 48.94 & 83.66 & 55.06 & 65.52 & 70.98 & 40.52 & 32.02 & 77.03 & 90.91 & 83.70 & 64.47 & 92.03 & 74.93 & 26.04  & 68.11 & 93.33 & 79.20\\

Gemma2-9B-Instruct\textsubscript{CR}   & 88.73 & \textbf{85.51} & 46.17 & 85.15 & 57.87 & 57.24 & \textbf{72.83} & 37.17 & 34.43  & 78.57 & 90.13 & 84.27 & 79.77 & \textbf{95.44} & 78.05 & \textbf{55.21}  & 76.64 & 96.56 & \textbf{81.88}\\

\hline

LLaMa3.1-8B-Instruct  & 84.29 & 81.80 & 47.02  & 81.23 & 56.74 & 63.45 & 66.71 & 31.60 & 39.23  & 74.71 & 89.21 & 79.43 &  68.83 & 84.71 & 77.20 & 19.79  & 71.03 & 87.78  & 77.26\\

LLaMa3.1-8B-Instruct\textsubscript{few-shot}   & 87.41 & 82.10 & 46.60 & 84.94 & 54.78 & \textbf{66.55} & 64.28 & 32.71 & 36.50  & 72.59 & 90.13 & 84.55 & 65.05 & 93.35 & 74.93 & 19.79 & 69.39 & 88.09 & 78.47\\

LLaMa3.1-8B-Instruct\textsubscript{CR} & 89.07 & 81.22 & 42.34 & 86.43 & 60.11 & 59.31 & 63.58 & 26.39 & 36.50 & 74.13 & 89.01 & 83.51 & 76.28 & 92.15 & 76.49 & 33.33   & \textbf{77.24} & 95.96  & 80.30\\

\hline

Qwen2-72B-Instruct  & 88.97 & 75.37 & 53.40  & 86.56 & 63.48 & 58.28 & 52.14 & 39.78 & 23.28 & 78.38 & \textbf{93.05} & 75.07  & 69.41 & 90.99 & 77.76 & 20.83   & 68.29 & 88.76 & 77.87\\

Qwen2-72B-Instruct\textsubscript{few-shot}  & 88.34 & 76.00 & \textbf{54.79} & 86.95 & 67.70 & 55.86 & 48.55 & 42.38 & 25.25 & 77.41 & 92.87 & 76.87 & 70.28 & 91.19 & \textbf{80.38} & 21.88  & 70.07 & 89.64 &  78.46\\

Qwen2-72B-Instruct\textsubscript{CR}   & \textbf{91.70} & 73.61 & 46.70 & \textbf{90.32} & \textbf{70.22} & 47.59 & 42.08 & 33.46 & 20.44 & 78.76 & 92.99 & 74.41 & 78.41 & 92.39 & 79.46 & 20.83  & 76.32 & 96.56 & 79.66\\

\hline

Mistral-7B-Instruct   & 82.82 & 83.46 & 48.51 &  79.44 & 67.98 & 61.72 & 54.10 & 33.09 & 24.59 &   71.43 & 91.06 & 78.67 & 65.05 & 90.63 & 70.68 &  18.75  & 64.10 & 85.00 & 75.66\\

Mistral-7B-Instruct\textsubscript{few-shot}  & 84.97 & 85.41 & 43.40 & 81.44 & 63.76 & 63.45 & 53.87 & 31.60 & 27.32 & 71.04 & 92.27 & 82.37 & 67.57 & 92.99 & 72.88 & 15.63   & 63.73 & 87.60 & 77.05\\

Mistral-7B-Instruct\textsubscript{CR} & 87.95 & 84.44 & 43.83 & 83.19 & \textbf{70.22} & 60.00 & 55.84 & 31.97 & 23.61  & 72.97 & 92.79 & 81.42 & 79.96 & 92.43 & 72.17 & 15.63  & 67.61 & 97.05  & 79.26\\
\hline

Random Sentence & 67.84 & 73.66 & 62.02 &  67.19 & 51.12 & 52.76 & 58.73 & 37.17 & 39.23 &   55.98 & 96.02 & 72.42   & 60.02 & 93.88 & 57.15 &  16.66   & 50.32 & 93.78  & 72.65 \\
\hline

\end{tabular}
}
\end{table*}

\subsection{Results and Analysis}
We analyze six LLM models on this benchmark dataset to evaluate their emoji recommendation and semantics preservation capabilities across the five downstream tasks, as shown in Table~\ref{tab:main_cls_results}. The results are obtained using three prompting approaches: zero-shot, few-shot, and conditional recommendation. All values are presented as percentages. 

It is worth noting that, \textit{in this context, if a model's F1 score is not particularly high, but its accuracy on the five downstream tasks is high, this indicates the model’s greater capacity for diverse emoji recommendation.} 

For instance, in the zero-shot setting, GPT-4o achieves the highest average accuracy at 79.23\%, although its F1 score of 9.04 ranks only third among the six models. This suggests that GPT-4o excels in recommending diverse emojis that, while differing from the ground truth, maintain the same semantic meaning.

In the few-shot approach, we observe performance improvements in most models except for GPT-4o. The Gemma2-9B-Instruct model demonstrates the best performance. The decline in GPT-4o's performance using the few-shot approach compared to the zero-shot approach could potentially be attributed to GPT-4o's inherently strong language understanding and reasoning capabilities. Providing examples in the prompt might constrain its emoji recommendation flexibility, limiting it to the scope of the five given examples.

In the conditional recommendation approach, we enhance the prompt by including user profile information such as gender and age when predicting emojis. This enhanced prompt significantly improves the performance of all models, with Gemma2-9B-Instruct achieving the highest average accuracy at 81.88\%. Across the board, models show an average 2\% improvement under this condition, highlighting that providing basic user profile information allows models to generate more accurate emoji recommendations tailored to users' needs.

This benchmark compares the performance of six existing models. We welcome the community to utilize our emoji benchmark to evaluate emoji recommendation and semantics preservation capabilities across a broader range of models.

\subsection{Case Study}
\label{sec: model_bias}
In this section, we present two key analyses that provide deeper insights into the performance of large language models on the semantics preserving emoji recommendation task.
One type of case study focuses on analyzing the classification accuracy of each specific class in each downstream task. Another case study examines the diversity capability of the models in emoji recommendation.

\subsubsection{Categorical Bias in Emoji Recommendation}

Table~\ref{tab:detail_cls} shows a detailed per-class accuracy on each task. We observe class bias in all the six LLM recommendation models.  For example, the accuracy of the female class in gender classification consistently surpasses that of the male class across all models. In the sentiment classification task, the neutral class shows noticeably lower accuracy compared to the positive and negative classes. A similar trend is observed in emotion classification, where models perform much better on the happiness class, while the disgust class exhibits poor performance. In the age classification task, the accuracy of the senior class is significantly lower than other classes,  with a value around 20\%.

 Recent studies~\cite{ferrara2023should} reveal that such categorical bias often arises when LLM models, faced with uncertainty in prediction, revert to heuristics and patterns ingrained from public datasets, resulting in predictions that conform to the most prevalent or recognizable trends within the training data. For instance, the bias towards the senior class may stem from the underlying assumption that most users posting on social platforms like Twitter are teenagers or middle-aged adults. Consequently, the models struggle to accurately classify posts from older users, often defaulting to the teen category. 

 Similarly, models tend to assume that users predominantly express happiness in their posts. When uncertain about user’s actual emotion, the models default to happiness, resulting in significantly higher classification accuracy for this class compared to other emotional categories.

\subsubsection{Diversity of Emoji Recommendation}

In this section, we evaluates how diverse the set of emojis recommended by different models is across the entire dataset. Table~\ref{tab:sorted_emoji_diversity_high_to_low} presents the unique emoji count for each model, sorted from the highest to the lowest. Among the models, GPT-4o stands out by recommending the largest number of unique emojis, while LLaMa3.1-8B exhibits the least diversity, with only 534 unique emojis. This variation in the number of unique emojis suggests that GPT-4o can potentially lead to more diverse emoji usage in real-world applications.

\begin{table}[ht]
\centering
\caption{Unique Emoji Count for Different Models, Sorted from High to Low.}
\begin{tabular}{l|c}
\toprule
\textbf{Model} & \textbf{Unique Emoji Count} \\ 
\midrule
GPT-4o    & {\bf 803}\\ 
Mistral-7B-Instruct   & 717 \\ 
Gemma2-9B-Instruct    & 664 \\ 
LLaMa3.1-70B-Instruct & 570 \\ 
Qwen2-72B-Instruct    & 554 \\ 
LLaMa3.1-8B-Instruct  & 534 \\ 
\bottomrule
\end{tabular}
\label{tab:sorted_emoji_diversity_high_to_low}
\end{table}

\begin{figure}[h]
    \centering
    \includegraphics[width=\linewidth]{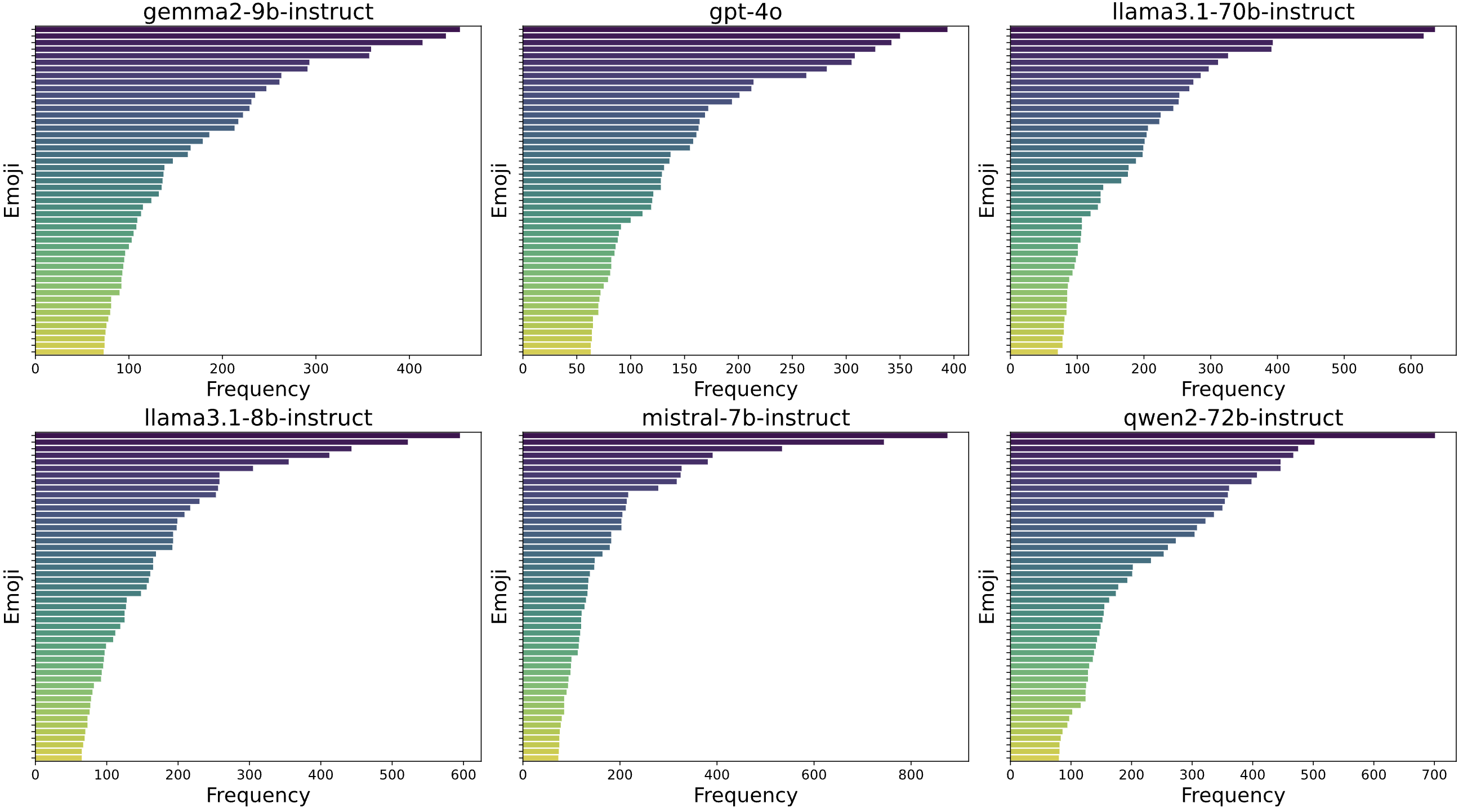}
    \caption{Distribution of Top 50 Frequently Used Emojis Across Different Models}
    \label{fig:emoji_frequency}
\end{figure}

Fig.~\ref{fig:emoji_frequency} complements this analysis by showing the distribution of the top 50 most frequently used emojis across six models. The distribution patterns reveal that models like Mistral-7B, Llama3.1-8B and Llama3.1-70B have a higher concentration on top 5 emojis, which are used much more frequently compared to others. On the other hand, GPT-4o-mini and Gemma2-9B display a more balanced distribution, suggesting that they recommend a wider range of emojis more equally.

\section{Conclusion and Future Work}
In this paper, we present a novel evaluation framework for semantics preserving emoji recommendation and  a comprehensive benchmark assessing the performance of various LLMs on this task. Our benchmark incorporates multiple evaluation metrics and investigates the impact of different prompting methods on model performance. We have demonstrated that Large Language Models, particularly GPT-4o, have significant potential in generating diverse and contextually appropriate emoji recommendations, especially when augmented with demographic information.

For future work, we plan to further enhance the benchmark by incorporating non-English corpora and expanding the dataset with more recent and diverse data from various platforms beyond X (formerly Twitter). This will improve the linguistic and cultural diversity of the benchmark. Additionally, the noticeable categorical bias of the existing LLM models analyzed in Section~\ref{sec: model_bias} motivates us to explore more accurate and unbiased emoji recommendation approaches, such as using Retrieval-Augmented Generation (RAG)~\cite{lewis2020retrieval} to obtain more unbiased data or fine-tuning debiased Large Language Models.

\bibliographystyle{IEEEtran}
\bibliography{references}

\end{document}